# Towards an Arabic-English Machine-Translation Based on Semantic Web

Neama Abdulaziz Bin-Dahan
*Department of Computer Science*
*Sana'a University, Sana'a, Yemen*
neama.abdulaziz@gmail.com

Fadl Mutaher Ba-Alwi
*Department of information systems*
*Sana'a University, Sana'a, Yemen*
fadlbaalwi@gmail.com

Ibrahim Ahmed Al-Baltah
*Department of Information Technology*
*Sana'a University, Sana'a, Yemen*
albalta2020@gmail.com

Ghaleb H. Al-gapheri
*Department of Computer Science*
*Sana'a University, Sana'a, Yemen*
drghalebh@yahoo.com

## Abstract

*Communication tools make the world like a small village, and as a consequence people can contact with others who are from different societies or who speak different languages. This communication cannot happen effectively without Machine-Translation because they can be found anytime and everywhere. There are a number of studies that have developed Machine-Translation for the English language with so many other languages, except the Arabic it hasn't been considered yet. Therefore, the aim of this paper is to highlight a roadmap for our proposed translation machine to provide an enhanced Arabic-English translation based on Semantic and to illustrate its work.*

## I. Introduction

Nowadays, technology makes our life much easier through the mitigation of the daily hardness. Different people from different societies can now communicate with each other easily and without needing to ask any other persons to be a translator during their business transactions or their conversations. Students can also study any foreign language, either online or by travelling to another country to get a certificate.

Languages mean culture that is why when we talk about Arabic language, we must talk about the Holy Quran. The Holy Quran is the richest Arabic document with the vocabularies, and Arab people consider it as the huge reference of keeping their language safe [1-3]. Therefore, people need to understand the Holy Quran to know as perfect as possible. However, Arabic language's structure is so complex. In addition, the vocabularies of the Arabic language depend on the derivation and morphology. English language, on the other hand, is much easier, and there are many people who can speak English like if they were native. It is also an essential subject in schools and there are many institutes and centers that teach it [4].

According to Al-Raheb, et al, [5] English language has just one rule to construct the sentence, i.e. subject (S) then verb (V) and after that the sentence complement (C) or the object (O). The sentence in the Arabic language, on the other hand, can have one of four different rules to construct the word, i.e. either S+V+(C,O), V+S+(C,O), V+(C,O)+S, (C,O)+S+V or S+(C,O)+V [5]. The homographs in both languages are also a huge problem while translation, because translation may give us another meaning that was not intended. Thus, knowing homographs is a very important task while translation [6]. Using a machine that can specify the intended meaning of the vocabularies used to build the sentence is highly needed.

We notice that, the semantic web provides us a strong means in order to build a translation machine. That translation machine can give us a more efficient translation than that provided in the online translator or any other types of translators or statistical machines. It is effective according to the coherent architecture that was developed to make analysis for the sentence more than once. First is to do a natural language analysis. Then to make morphological analysis beside the context analysis to know the real meaning that the source means. After that is to use the statistical analysis to get the final translation that the machine will send to the target. This issue still found in the Arabic language, which is probably the main reason for the absence of complete and efficient translation machines in the state of the art. This is in turn provided an opportunity for researchers to enrich this topic further. From another point of view, most of current researches have focused on the parser technique only. Because of that, this study intends to do a semantic analysis in addition to the statistical analysis. The semantic analysis can be divided into a morphological analysis and ontology to specify the correct meaning in the other language.

The researchers demonstrate that the morphological analysis should be done first before the translation. Then, using the normal statistical Machine Translation (MT) could be quietly enough to get the right meaning of the sentence.

The aim of this paper is to propose a translation machine to provide an Arabic-English translation. The reminder of this paper is organized as follow: Section 2 presents the related works in some details. The proposed machine model is demonstrated in section 3. Section 4 presents the discussion of this study. Section 5 concludes this study.

## II. Related work

Today's machine translations (MTs) have many different parts to give us the most efficient translation it can. Thus, there are so many sets of research that contribute to developing these parts. We are going to talk here about some developments in MTs, parsers and Ontologies.

### 1. Machine-Translation (MT)

The statistical method was introduced to be a new approach for MT [7]. Statistical MT was stated by IBM researchers in [8]. They thought that MT is old as the first generation of the computers. They also stated that the translation must depend on many factors, the most essential one was the whole original text itself. In contrast, they treated the words without recognizing the connection between words or even recognizing the sentence structure. Hutchins stated that the MT system may be designed to satisfy the following criteria: to deal with single words, to get restricted input text structure, to have pre-edited input texts with any grammatical category, and without caring about the ambiguity of words or any other operators during translation. They also classified the MT to many groups according to the type of aiding (machine-aided MT, human-aided MT, computer-aided translation), input or output edition (pre-edited or post-edited), number of targeted languages (bilingual or multilingual), translation approach (direct translation, interlingua, or transfer) and syntactic structure analysis (predictive analysis, phrase structure or dependency grammar). They also pointed to the importance of semantic MT, but it was just a survey and there were no implementations for their vision [9]. Chan, et al [10], integrated a state-of-the-art Word Sense Disambiguation system into a state-of-the-art hierarchical phrase-based MT, "Hiero" [10]. In contrast, they demonstrate only one way for the integration without introducing any rules that compete against existing rules [11]. Gupta et al, [11] introduced 16 features that were extracted from the input sentence and their translation and then showed a quality score based on Bayesian inference produced from their sample training data, but they didn't develop a new English-Hindi MT and their work was just an analysis study for an existing MT [11].

### 2. Arabic Parsers

Natural Language parser is a machine that can understand the sentence parts and serve us in the translation using the MT. There were many researchers who have worked on this type of machine, but, here, we are going to explore some of the most modern Arabic Parsers which made a real differentiation in Arabic translation and in Arabic natural language processing.

An example of the parsers is Microsoft ATKS as

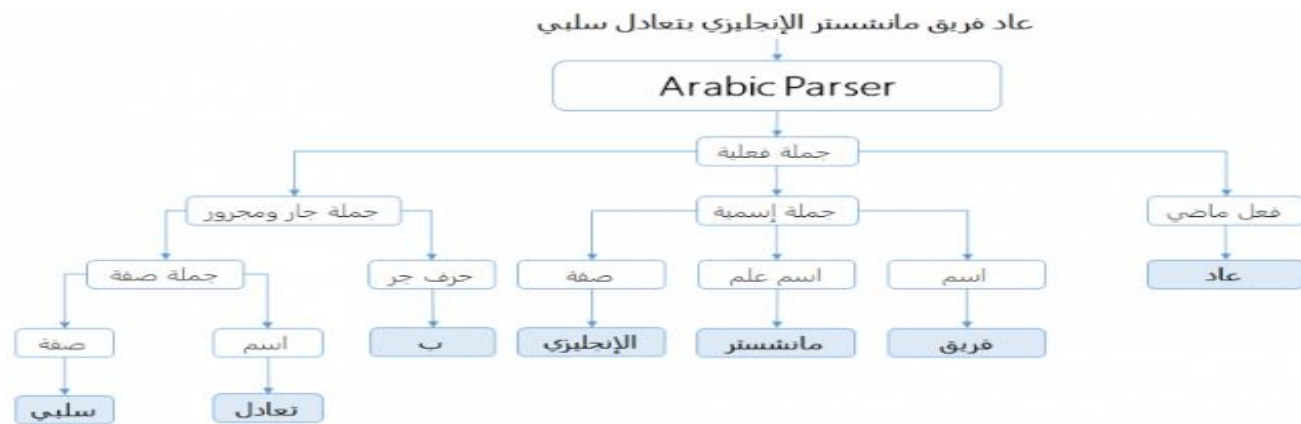

**Figure 1: Microsoft's ATKS Parser analysis for Arabic sentence**

shown in figure 1. This parser was used by Alagha, et al in [1-3] to simplify the process of converting the Arabic sentence into the RDF triples. Arabic Parser was proposed in [12-15]. English language has just one rule to construct the sentence, i.e. subject (S) then verb (V) and after that the sentence complement or object (C,O). However, Al-Raheb, et al [5], stated that, Arabic is a free word order: SV(C,O), VS(C,O), V(C,O)S [5]. In our vision, the structure of the Arabic sentence can be more various than that they stated. In addition, according to Green, et al [16], their parser was similar to another Treebank in gross statistical terms, annotation consistency remains problematic [16]. Tounsi, et al [17], thought that the best-known Arabic statistical par ser, at that time, was developed by [17]. Thus they tried to enrich that parser's output with more abstract and deep dependency information. On the other hand, the categories they added to Bikel's parser resulted in substantial data-sparseness

[18]. Green, et al [16], introduced the Stanford gold parser and showed that, in their paper, Arabic parsers is poorer in thoughts and still much lower than English ones. Their proposed parser, according to Stanford University, is demonstrated as one of the most effective and found parsers [16]. Al-agha, et al [1], developed a new parser which has many functionalities over than others [1]. Zaghouani, et al [19], added new specified post-editing rules of correction, according to seven categories: spelling errors, word choice errors, morphology errors, syntactic errors, proper name errors, dialectal usage errors, and punctuation errors [19].

## 3. Ontology-Based Machine-Translation

all of them can only serve for their proposed languages. They also mentioned that the instant translators like google, bing and more others are statistical translators and have no semantic processing in their works [6]. Alagha, et al [1], presented a domain-independent approach to translate Arabic NL queries to SPARQL by getting benefits from linguistic analysis [1]. It was just to translate the Arabic questions to SPARQL, not to other human-language. The systems developed in [2,3] take questions expressed in Arabic and returns the answers drawn from an ontology-based knowledge base though the ontology file was not prepared as the system needs. While the translation is done to the SPARQL level only, it would not work as an MT between human-languages and it would be like the previous works that didn't translate

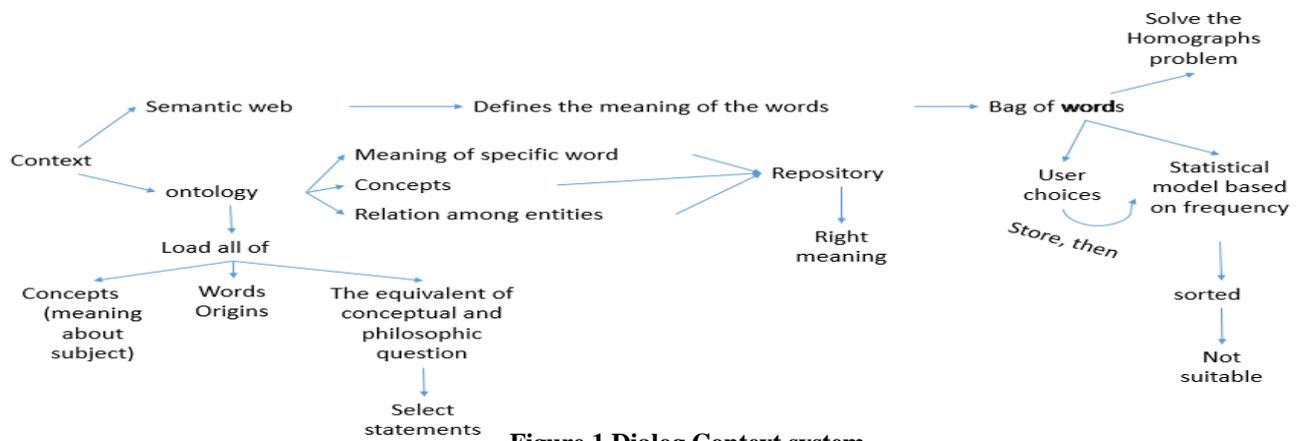

Figure 1 Dialog Context system

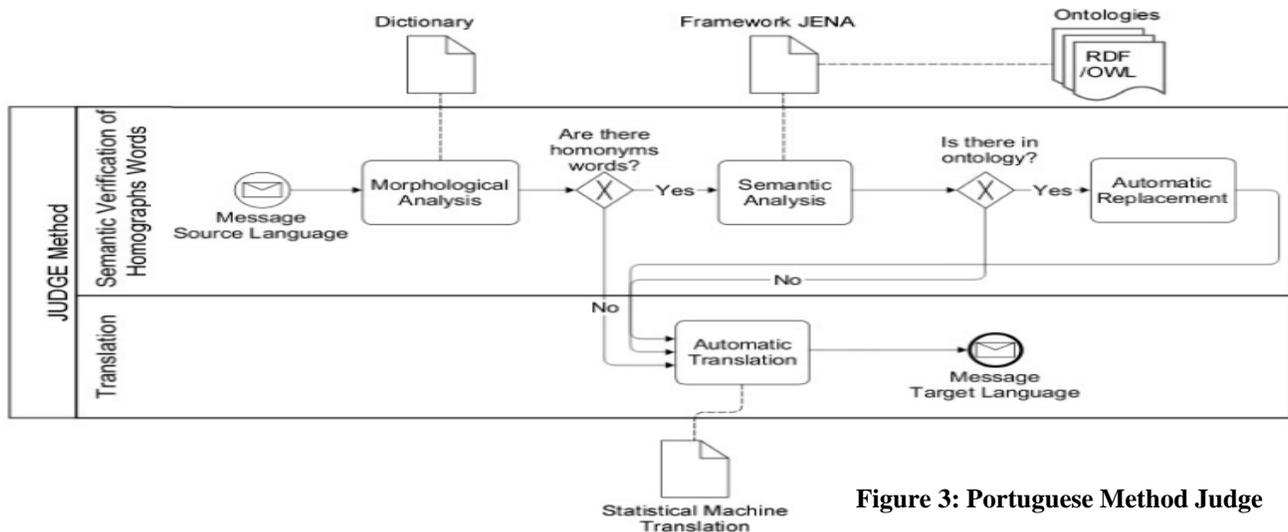

Figure 3: Portuguese Method Judge

Shi, et al, developed a world model for Chinese-English MT using an Ontology-driven [20]. Seo, et al [21], presented a syntactic and semantic method for English-Korean MT using Ontology for web-based MT [21]. Mousallem, et al [6], solved the ambiguity in dialog conversation using an ontology-based MT [6]. However, the Arabic question sentences to any other human-languages.

The main solution of the Portuguese-English MT was to provide an efficient way to process the homographs [6]. And where the last dialog systems were depending on the statistical MTs. Thus, they proposed a semantic analysis before the traditional translation of the statistical MT. Figure 2 shows the mental plan, from which their proposed model was invented. Then they developed a model which was named the Method Judge as shown in figure 3. It consists of two essential parts. The semantic verification is to show the right meaning of the ambiguous words or the homographs which is followed by the automatic replacement of the vocabularies meanings. There will be a morphological analysis connected with the source language dictionary to specify the right meaning of the homograph words. There is also a semantic analysis to get the right meaning of that homograph from the language ontology. Then the resulted words will be replaced automatically with their effective meanings. The second part was the automatic translation of the remaining parts of the sentence sent in the dialog system. This type of translation will be done statistically using a cluster algorithm, like the way used in the instant translators such as google and bing translators.

According to Mousallem, et al [6], both languages have the same order of the sentence parts, there was no need to have two different parsers to process the natural language. The parser, in that research, was merged in the different online translators (Google, bing, Worldling, Gengo and Systran) and the results were very interesting [6].

Al-agha, et al [1-3], proposed a machine to translate the Arabic questions to SPARQL. They used the ontology for the linguistic and the semantic analysis and to eliminate the ambiguity of homographs [3], although the system does not make intensive use of sophisticated linguistic methods. They also defined a Natural Language (NL) interface for questions formulated in natural language to return answers on the basis of a given which was that those systems were not able to retrieve precise answers to questions, but only to retrieve a set of relevant documents using the given keyword-based query [1-2].

## III. The proposed Machine Model

We can extend the Method Judge mentioned in [6] with the parsers idea in a model that can be illustrated in figure 4. The Main method judge was explained in the last point. The extended MT system will be built from many morphological analysis model, in which the sentence will be divided into its main parts and they stated here that the homographs may be either a noun or a verb. Thus, the automatic replacement will never generate any error. Even though the auxiliary verbs in one of the proposed languages (English or Portuguese) has a different meaning or usage, the meaning will be efficiently specified because they both have the same sentence structure. The researchers test their model with five other toolkits. The model assumes that, after choosing the source language, the machine must be able to detect the language from context before the source sentences are passed through a parser. The parser will specify the parts-of-speech in that language. We can after that decide which one of these words can be considered as a homograph. The homographs will be analyzed using the morphological analysis. Using the language ontology and the context of the source text, the correct translation will be chosen. The semantic analysis will be used to make sure which meaning of the available meanings of that homograph word is the most closer to the context. After that, a simple statistical MT will be done. Here must be a toolkit that must make sure the precision of the translation of the target language. Thus, we propose that toolkit will be the parser of the target language.

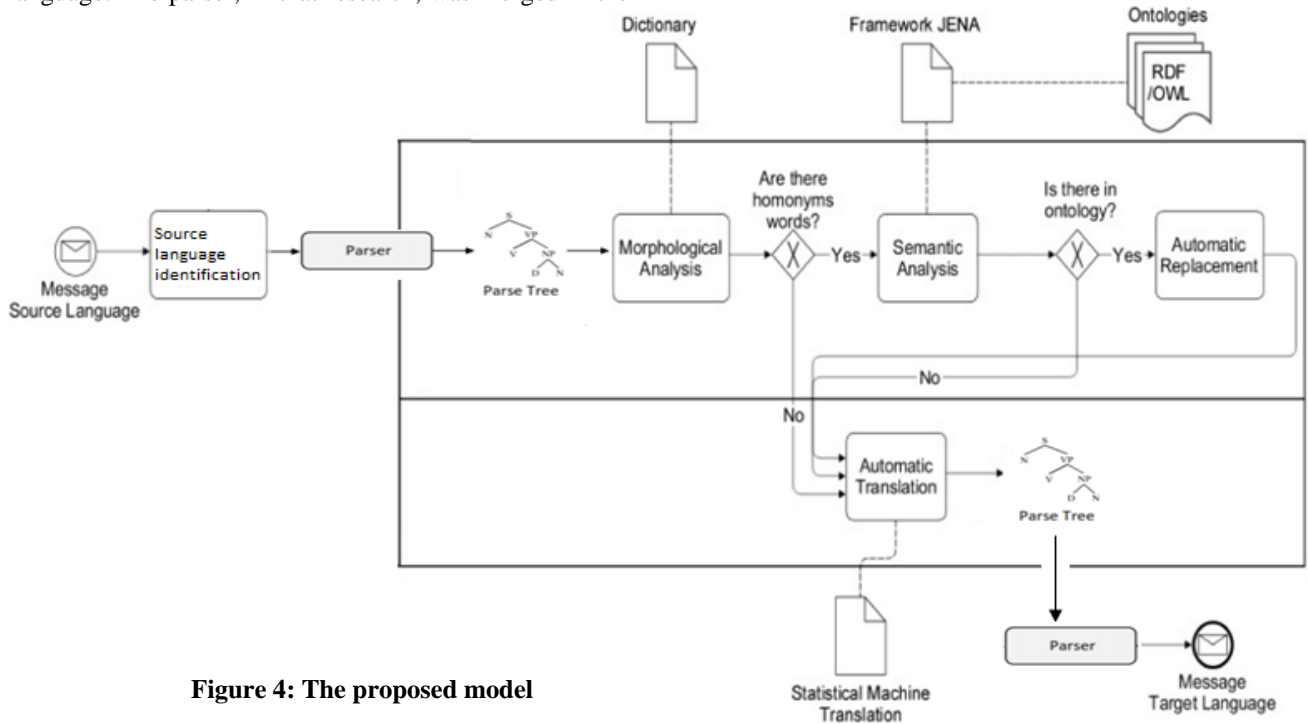

**Figure 4: The proposed model**

## IV. Discussion

The extended model will adopt the Method Judge as the heart of the MT. It must ensure the precision of the translation between the Arabic language and English language according to two standards: the precision of the sentence grammar and structure, and the precision of the words meanings which the MT writes during the translation.

Naturally, the Arabic Parser is different from the English one. Therefore, we assume that there must be a source language identification process that can be triggered automatically even though the user didn't till the machine what is the source language. The parser of the source must not be the parser of the target. Then the source language sentences will be passed through a Natural language processing toolkit. After that, the Method Judge must start working to give us the precise translation according to its morphological and semantic analysis. One of the included ontologies at a time will be chosen according to the source language. The other may not be used because the rest of the translation can be made by the statistical translation effectively. Once it is ready, the aggregated result will be sent to the target language parser to correct the structure of the resulted sentence before showing the final result to the end user.

Like the translation in all the MT papers which was proven mathematically and the results were measured using the F-measure, the results of the proposed machine must be tested using that measure, which can be calculated as the following [1]:

$$\text{Precision} = \frac{number\ of\ correctly\ translated\ queries}{number\ of\ queries\ generated\ by\ the\ system}$$

$$\text{Recall} = \frac{number\ of\ correctly\ translated\ queries}{number\ of\ testing\ queries}$$

$$\text{F-measure} = \frac{2 \times Precision \times Recall}{Precision + Recall}$$

## V. Conclusion

Semantic Web technologies help us to solve many problems in trading, learning and communication. Building a statistical Machine-Translation was inefficient to give us the correct and right meaning of the sentence and its words in the context. Semantic Analysis with all of its partitions gives us a perfect solution to get a better translation in comparison with some statistical ones.

There were many successive semantic MTs that translate between English language and many other languages. The ones which depend on specifying the meaning of the homographs from the context were much more powerful. We are trying to develop a semantic MT between Arabic and English languages using the facilities provided by the semantic web. This machine will depend on the latest extended Arabic ontology, or set of ontologies, and also an English one to provide an instant translation from Arabic to English and vice versa. In this research we highlight a proposed model for a MT. We also assumed that this MT will provide an enhanced translation in comparing with the results of the existent instant translators.

As a future work, if the ontologies are not as comprehensive as needed, we suggest making an extension from all the available ontologies and to add more terms if there's still a need. We also suggest develop this MT to support more languages. The MT must also be optimized to translate long texts and Arabic artistic texts.

## References

[1]. I. AlAgha, and A. Abu-Taha. "AR2SPARQL: An Arabic Natural Language Interface for the Semantic Web." *International Journal of Computer Applications*, 125.6, 2015

[2]. I. AlAgha, and A. Abu-Taha. "An Ontology-Based Arabic Question Answering System" *Central library of Islamic Univeristy of Ghaza*, 2015

[3]. I. AlAgha.. "Using Linguistic Analysis to Translate Arabic Natural Language Queries to SPARQL." *arXiv preprint arXiv*:1508.01447, 2015

[4]. A. Alqudsi, N. Omar, and K. Shaker. "Arabic machine translation: a survey." *Artificial Intelligence Review,* 42(4), 2014, pp. 549-572.

[5]. Y. Al-Raheb, A. Akrout, J. van Genabith, J. Dichy. "DCU 250 Arabic Dependency Bank: An LFG Gold Standard Resource for the Arabic Penn Treebank" The Challenge of Arabic for NLP/MT at the British Computer Society, UK, 2006, pp. 105-116.

[6]. D. Moussallem, and C. Ricardo. "Using Ontology-Based Context in the Portuguese-English Translation of Homographs in Textual Dialogues." *arXiv preprint arXiv*:1510.01886, 2015.

[7]. W. Weaver. "Translation." *Machine translation of languages,* 14,1955, pp. 15-23.

[8]. P. F. Brown, J. Cocke, S. A. D. Pietra, V. J. D. Pietra, F. Jelinek, J.D. Lafferty, and P. S. Roossin. "A statistical approach to machine translation." *Computational linguistics,* 16(2), 1990, pp. 79-85.

[9]. J. Hutchins. "Machine translation: History and general principles." *The encyclopedia of languages and linguistics*, 5, 1994, pp. 2322-2332.

[10]. Y. S. Chan, H. T. Ng, and D. Chiang. "Word sense disambiguation improves statistical machine translation." *Annual Meeting-Association for Computational Linguistics* 45(1), 2007, p. 33.

[11]. R. Gupta, N. Joshi, and I. Mathur. "Analysing quality of english-hindi machine translation engine outputs using Bayesian classification." *International Journal of Artificial Intelligence and Applications*, 4(4), 2013, pp. 165-171.

[12]. E. Ditters. "A formal grammar for the description of sentence structure in modern standard Arabic." *EACL 2001 Workshop Proceedings on Arabic Language Processing: Status and Prospects*. 2001.

[13]. E. Othman, K. Shaalan, and A. Rafea. "A chart parser for analyzing modern standard Arabic sentence." *Proceedings of the MT summit IX workshop on machine translation for semitic languages: issues and approaches*. USA, 2003

[14]. A. Ramsay, and H. Mansour. "Towards including prosody in a text-to-speech system for modern standard Arabic". *Computer Speech and Language*, 22, 2007, pp. 84-103.

[15]. Z. Zabokrtsky, and O. Smrz. "Arabic syntactic trees: from constituency to dependency". *Proceedings of the tenth conference on European chapter of the Association for Computational Linguistics,* European Chapter of the Association for Computational Linguistics EACL 2003, pp. 183-186.

[16]. S. Green, and C. D. Manning. "Better Arabic parsing: Baselines, evaluations, and analysis." *Proceedings of the 23rd International Conference on Computational Linguistics*, Association for Computational Linguistics, 2010, pp. 394-402

[17]. D. M. Bikel. "Intricacies of Collins' parsing model." *Computational Linguistics*, 30(4), 2004, 479-511.

[18]. L. Tounsi, M. Attia, and J. van Genabith. "Parsing Arabic using treebank-based lfg resources." 2009.

[19]. W. Zaghouani, N. Habash, O. Obeid, B. Mohit, H. Bouamor, and K. Oflazer. "Building an Arabic machine translation post-edited corpus: Guidelines and annotation", *International Conference on Language Resources and Evaluation (LREC 2016).* 2016.

[20]. C. Shi, and H. Wang. "Research on ontology-driven Chinese-English machine translation." *2005 International Conference on Natural Language Processing and Knowledge Engineering*. IEEE, 2005

[21]. E. Seo, I. S. Song, S. K. Kim, and H. J. Choi, "Syntactic and semantic EnglishKorean machine translation using ontology." *Advanced Communication Technology, 2009. ICACT 2009. 11th International Conference on*. IEEE, 3, 2009.